\documentclass[letterpaper,10pt,conference]{ieeeconf}
\IEEEoverridecommandlockouts                              

\usepackage{graphicx} 
\graphicspath{{static_figs/}{generated_figs/}}
\usepackage{times} 
\usepackage{amsmath} 
\usepackage{amssymb}  
\usepackage{lipsum}
\usepackage{fancyhdr}
\usepackage{algorithmic}
\usepackage{algorithm}
\usepackage{booktabs}
\usepackage{flushend}

\title{%
  \LARGE \bf
  Acoustic Beamforming for Object-relative Distance Estimation and Control
  in Unmanned Air Vehicles using Propulsion System Noise
}

\author{%
  Alisha Sharma$^{1,2}$,
  Jason Geder$^{1}$,
  Joseph Lingevitch$^{1}$,
  Theodore Martin$^{1,3}$,
  Daniel Lofaro$^{1}$, and
  Donald Sofge$^{1}$,
\thanks{$^{1}$%
        Naval Research Laboratory,
        Washington, DC, 20375, USA
        }%
\thanks{$^{2}$%
        University of Maryland, College Park
        College Park, MD, 20742, USA
        }%
\thanks{$^{3}$%
        Excet Inc.,
        Springfield, VA, 22150, USA
        }%
\thanks{This work was funded by an Office of Naval Research program.}
\thanks{Distribution Statement A: Approved for public release. Distribution is unlimited.}%
}

\begin{document}

\maketitle
\thispagestyle{fancy}

\begin{abstract}

  Unmanned air vehicles often produce significant noise from their propulsion systems. Using this broadband signal as “acoustic illumination” for an auxiliary sensing system could make vehicles more robust at a minimal cost. We present an acoustic beamforming-based algorithm that estimates object-relative distance with a small two-microphone array using the generated propulsion system noise of a vehicle. We demonstrate this approach in several closed-loop distance feedback control tests with a mounted quad-rotor vehicle in a noisy environment and show accurate object-relative distance estimates more than 2x further than the baseline channel-based approach. We conclude that this approach is robust to several practical vehicle and noise situations and shows promise for use in more complex operating environments.

\end{abstract}

\section{Introduction}\label{sec:intro}

Autonomous unmanned air vehicles (UAVs) rely on a range of sensing techniques to navigate, including GPS, lidar, and cameras; however, all sensing techniques fail in certain operating conditions and sensors can be damaged in operation.
Many UAVs produce significant broadband noise from their motor-propeller systems.
Harnessing components of this noise for large-object distance estimation in an auxiliary passive sensing system could make UAV navigation more robust with minimal cost.

This task is challenging for several reasons.
First, the problem is dominated by noise as onboard microphones are much closer to the propellers than the objects we are trying to sense.
To make matters worse, while explicit chirps used in active sonar are designed for performance, motor-propeller noise is highly random and complex.
Finally, because this system is meant to be mounted on a UAV, we have significant weight, power, and cost constraints that make many standard approaches impractical.

Several approaches have been proposed for low-resource, hardware-agnostic acoustic sensing with small arrays, including leveraging sensor motion \cite{Cur91}, incorporating physical structures for directionality \cite{Ma16,Bai22}, and location and source-aware sensing \cite{Ogi15,Sha22}, among others.
Recently, \cite{Cal20,Cal21} studied the feasibility of using propulsor self-noise to estimate the distance from a vehicle to a rigid reflector for simplified scenarios involving a single propeller and a single \cite{Cal20} or two \cite{Cal21} microphones.
We hypothesize that cross-correlating acoustic beams (``look directions'') steered at the reflecting wall and the propellers may provide a higher signal-to-noise ratio than the channel approach in \cite{Cal20}, enabling control in more challenging conditions.

\begin{figure}
  \centering
  \includegraphics[width=\linewidth]{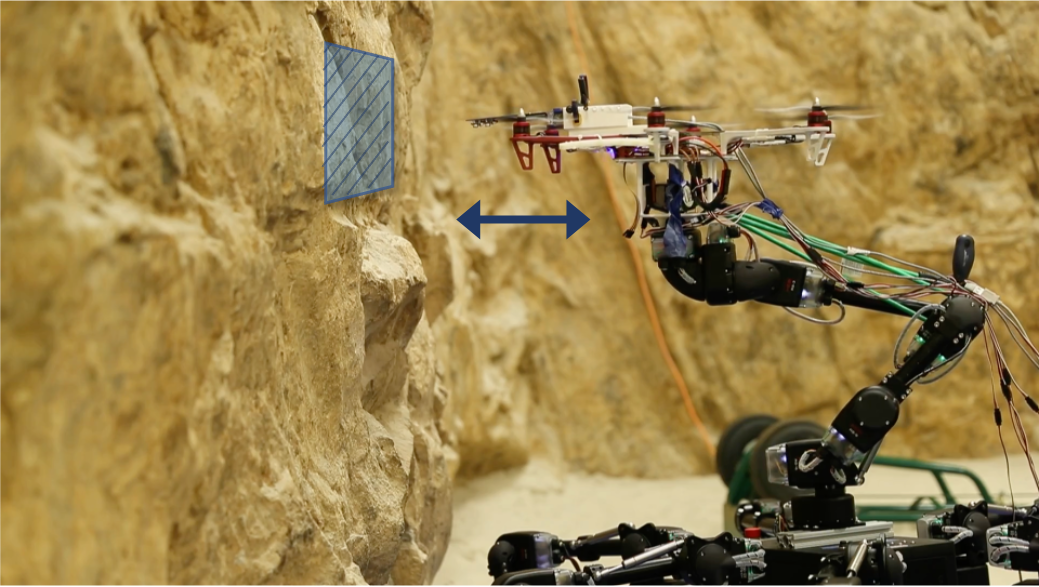}
  \caption{We use the noise generated by a vehicle's propulsion system for object-relative distance estimation and feedback control. In this work, we focus on sensing using a mounted quad-rotor UAV. This photo shows our test setup, in which UAV is mounted to a 6DoF actuating arm to approximate free flight.}
  \label{fig:arm}
\end{figure}

In this work, we present an effective beamforming-based distance estimation algorithm and demonstrate it in active control of a practical quad-rotor UAV.
Our major contributions are as follows:
\begin{enumerate}
  \item We introduce an \textbf{acoustic beamforming object-relative distance estimation algorithm} using two microphones from the implicitly generated propulsion system noise.
    This beamforming approach consistently outperforms the channel autocorrelation difference algorithm \cite{Cal21} in simulation and several open- and closed-loop distance estimation experiments. Furthermore, this approach can be easily extended in the future for orientation estimation and larger arrays.
  \item We \textbf{implement this code fully onboard the vehicle} using a Teensy 4.0 microcontroller. We use this to make real-time acoustic distance estimates.
  \item We \textbf{demonstrate closed-loop feedback control} of a mounted quad-rotor UAV in a noisy ambient setting using this beamforming-based algorithm.
  \item Finally, we show that the two-microphone acoustic beamforming algorithm can \textbf{estimate object-relative distance up to nearly 0.5 m outboard with a useful accuracy}, more than a 2x improvement on the channel-based method.
\end{enumerate}

\begin{figure}[h!]
  \centering
  \includegraphics[height=1.8in]{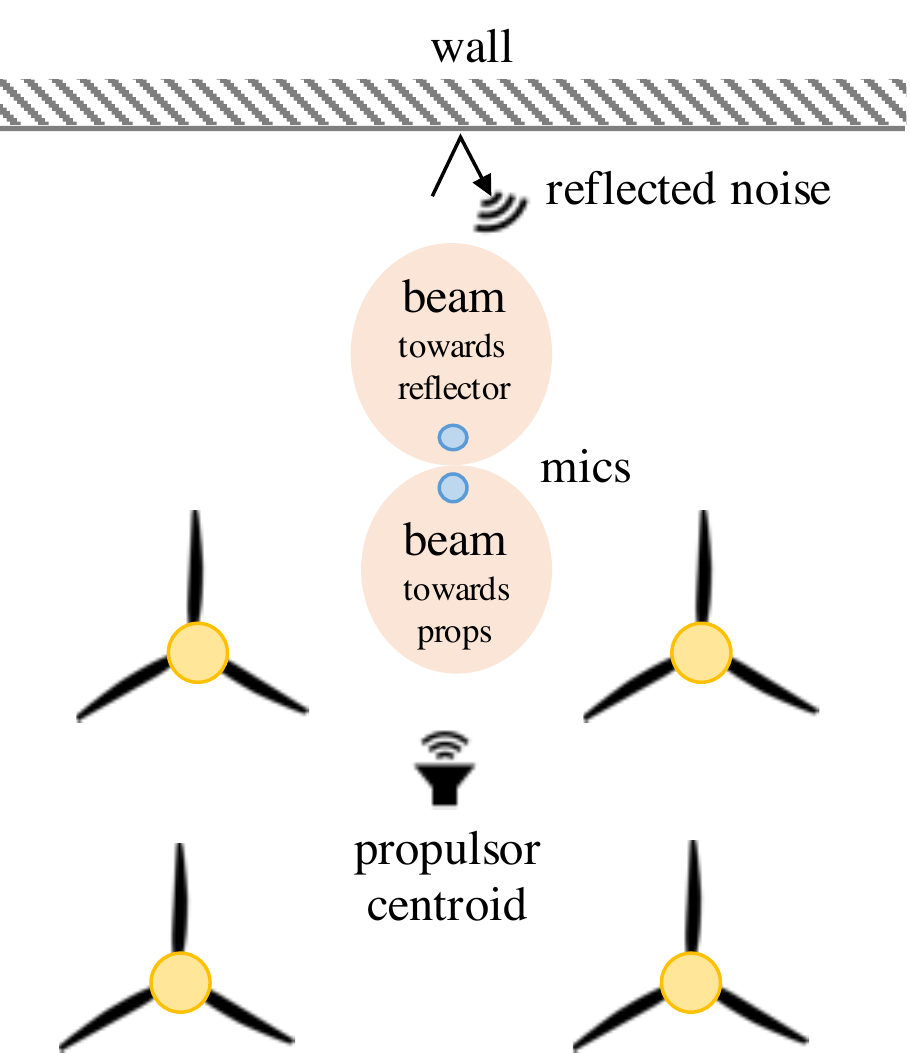}
  \hspace{0.1in}
  \includegraphics[height=1.8in]{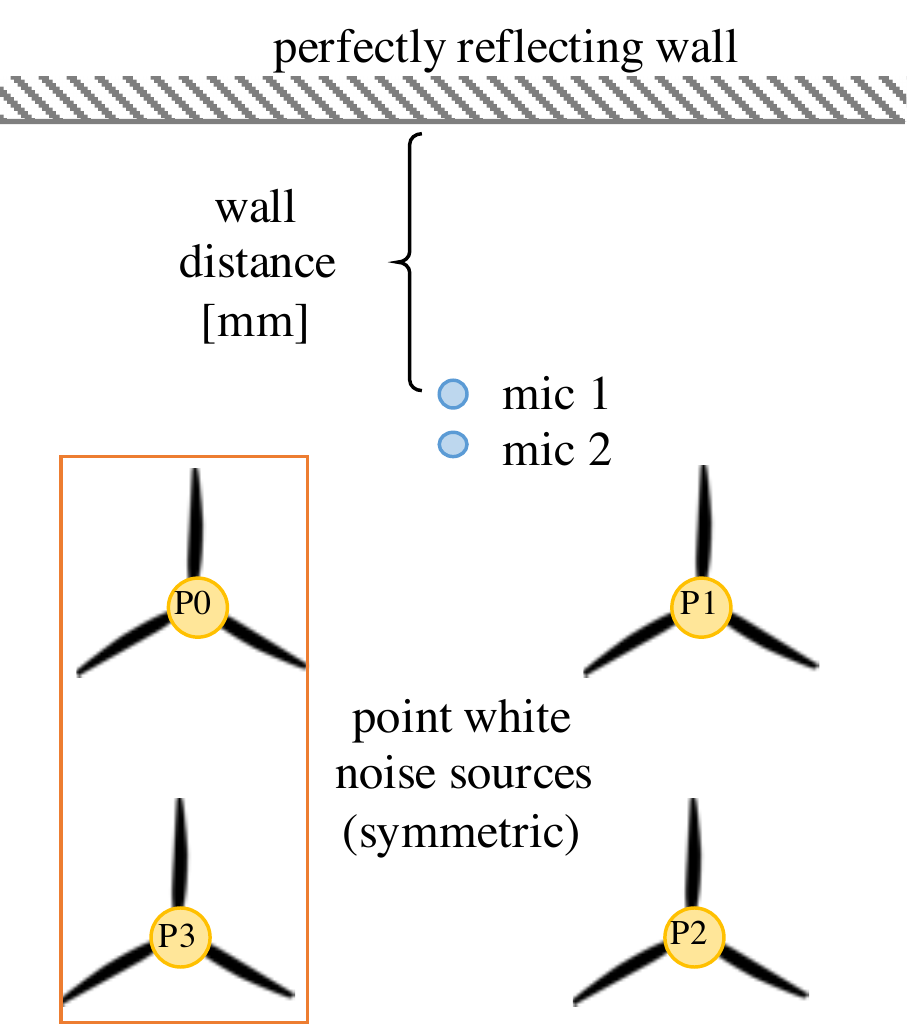}
  \caption{%
    (Left) Beamforming conceptual approach. Using a two-mic array, steering vectors are constructed pointing towards the wall (for the reflected signal) and the propulsors (for reference). These beams are then correlated to estimate the time delay of the reflected signal, and by extension, object-relative distance of the array.
    (Right) simulation geometry, which mimics the quad-rotor experimental geometry in later sections. We show results for sources P0 and P3, indicated by the orange box.
    }
  \label{fig:theory-setup}
\end{figure}

\section{Problem Definition}\label{sec:definition}

Previous work \cite{Cal20,Cal21} studied the feasibility of using propulsor self-noise to estimate the distance from a vehicle to a rigid reflector for simplified scenarios involving a single propeller and a single \cite{Cal20} or two \cite{Cal21} microphones.
The interference between the direct path and reflected noise is observed as intensity variation in the frequency domain that depends on the path length difference and acoustic frequency.
For example, the path difference can be observed by calculating the autocorrelation spectra of each microphone’s response \cite{Cal21}.
The interference pattern is straightforward to discern for an idealized white noise source due to an equal contribution for each frequency in correlation analysis.
However for motor/propeller systems the acoustic signal is not white and contains strong frequency lines at frequencies associated with the propeller harmonics, as well as components from the motor and electronic speed controller, which can dominate correlation methods and hide the path length interference.

In \cite{Cal21}, a time-domain approach using a comparison between the autocorrelation spectra of two microphone channels was implemented with a feedback loop to the propulsor to achieve a vertical separation between a negatively buoyant blimp and the floor.
In this approach, the Fourier spectra of each microphone was normalized to reduce the dominance of the propellor and motor harmonics.
The normalization factor was the average magnitude over both microphones in each Fourier bin, which serves to preferentially weight the spectral bins as a comparator while reducing the influence of the harmonics.

In this work, we generalize the approach to problems involving multiple motor-propellers (for example, in a multi-rotor UAV) and high ambient noise conditions.
In contrast to prior approaches, we implement a beamforming cross-correlation analysis using arrays of two or more microphones.
We incorporate the Fourier bin normalization methodology from \cite{Cal21} to prevent the motor-propeller harmonics from overwhelming the analysis.

\subsection{Acoustic Beamforming Approach}

\begin{figure}[ht]
  \centering
  \includegraphics[height=1in]{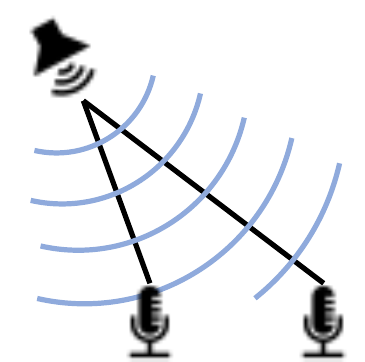}
  \caption{Illustration of the phase difference in different microphones of a beamforming array.}
  \label{fig:beamforming}
\end{figure}

\textit{Acoustic beamforming} is a spatial sensing technique that uses sensor arrays to estimate signal directionality \cite{Opp99}.
This is done by designing arrays such that direction can be inferred from the phase difference of the incoming signal (Figure \ref{fig:beamforming}),
which allows us to steer \textit{beams} in specified directions (``look directions'').
In this work, we focus on spatial filtering with small two-microphone arrays to improve distance estimation.

To account for multiple noise sources, we generalize the channel-based correlation approach
described in \cite{Cal21}
to utilize acoustic beamforming (Algorithm \ref{alg:bf}).
Here we consider beamforming approach applied to a simple microphone array (2 microphones);
more complex arrays will achieve better beamforming results at the expense of increased hardware and signal processing requirements.

\begin{algorithm}
  \begin{algorithmic}
    \STATE \textbf{Require:} look directions $\mu_1,\mu_2$
    \WHILE {system is running}
      \STATE Read raw audio buffers from each mic
      \STATE Apply Hann window
      \STATE Compute FFTs of windowed buffers
      \STATE Compute moving averages of cross-spectral density (CSD) of FFTs
      \STATE Average CSD over moving average arrays
      \STATE Normalize CSD by trace
      \STATE Form steering vectors with $\mu_1,\mu_2$
      \STATE Apply beamforming to get beam power spectrum
      \STATE Window, normalize beam power spectrum
      \STATE time delay $t_d\gets$ IFFT(beam power spectrum)
    \ENDWHILE
  \end{algorithmic}
  \caption{Beam cross-correlation}
  \label{alg:bf}
\end{algorithm}

The basic idea is to cross correlate beams steered away from the vehicle with a reference beam steered at a propeller noise source
(Figure \ref{fig:theory-setup}, left);
a correlation peak is observed for a beam steered in the direction of the reflected ray path. Pseudocode for this approach can be seen in Algorithm \ref{alg:bf}.

\subsection{Simulation Results}

\begin{figure}
  \centering
  \includegraphics[width=\linewidth]{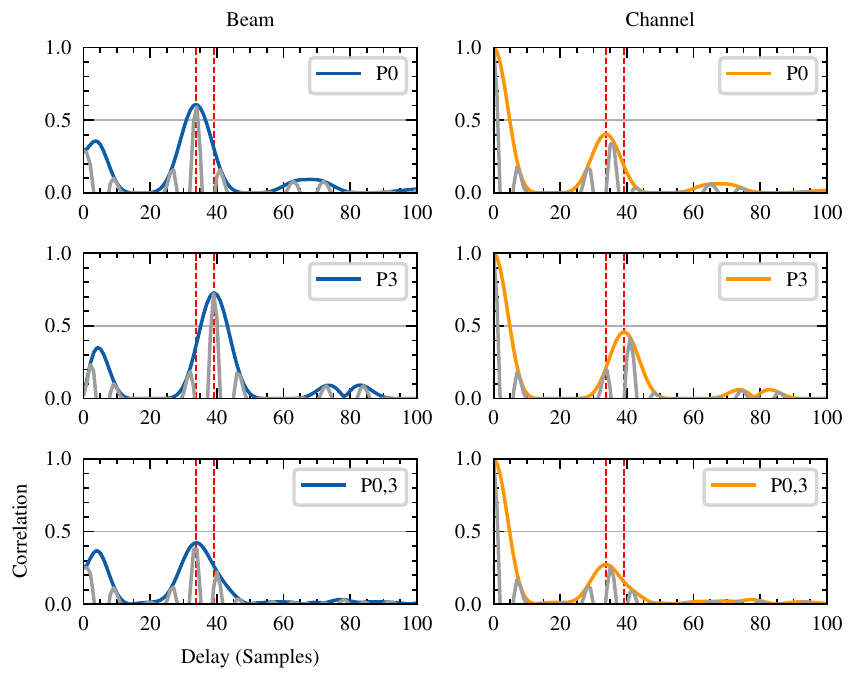}
  \caption{Beam cross-correlation (left) compared with channel autocorrelation difference (right) for a two-source system with no ambient noise.
  The top row shows response for source P0 only,
  the middle shows P3 only,
  and the bottom shows results for both P0 and P3.
  The gray curves show the raw correlation, and the colored curves are the complex envelope computed with a Hilbert transform of the correlation.}
  \label{fig:sim}
\end{figure}

In this section, we compare channel-based autocorrelation difference and beam-based cross-correlation in a simulated symmetric multi-rotor setting.
The approximate geometry is shown in Figure \ref{fig:theory-setup} (right),
with sources P0 and P3 at $(-0.1375, -0.255)$ m and $(-.1375, -.731314)$ m respectively.
Noise sources are synthetically generated white noise from uncorrelated noise sources in a 2D geometry.
The wall reflector is positioned at $y=0$, and sound speed is constant at 343 m/s.
We use a 2-element microphone array (15 mm separation.) Individual microphones are positioned at $(0, -0.1475)$, $(0, -0.1625)$ m.

Simulation results can be seen in Figure \ref{fig:sim}.
The columns denote algorithm (left: beam-based, right: channel-based), and the rows denote noise sources (top: P0 only,
middle: P3 only,
bottom: P0 + P3).

We make two primary observations.

\begin{enumerate}
  \item First, peaks from both beam- and channel-based methods attenuate in the multi-source case (Figure \ref{fig:sim}, bottom row); however, this attenuation is more dramatic in the channel-based case due to the already low amplitude of the signal.
  \item Second, the beamforming correlation (Figure \ref{fig:sim}, left) is much higher-amplitude than the corresponding channel-based signal (Figure \ref{fig:sim}, right), which should make it easier to identify in a distance estimation algorithm, particularly in high-ambient-noise conditions (better signal-to-noise ratio).
\end{enumerate}

From these simulations, we expect that a beam-based correlation approach will be more robust to multi-rotor settings with high ambient noise.
In Section \ref{sec:results}, we test this hypothesis in an experimental setting.

\section{Methods}\label{sec:methods}

The following experiments use a quad-rotor UAV mounted on an actuating arm to demonstrate feedback control of wall-relative distance in a constrained experimental setup prior to free-flight testing.
We estimate wall distance using algorithms that take in sound data from an array of two mounted microphones.
All distance estimates are computed in real time using an onboard microcontroller.

During open-loop control experiments, the actuating arm moves the vehicle in pre-defined patterns, and we log the raw audio signal, onboard distance estimates, and ground truth distance for each run.
During closed-loop control experiments, distance estimates are computed onboard in real-time then are sent to a proportional controller on the actuating arm that responds to differences between commanded and measured wall distance.

\subsection{Vehicle Setup}\label{sec:vehicle}

\begin{figure}[ht]
  \centering
  \includegraphics[width=\linewidth]{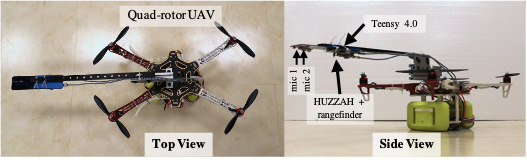}
  \caption{Top (left) and side (right) views of the vehicle geometry.}
  \label{fig:uav}
\end{figure}

Our vehicle is a quad-rotor UAV, as seen in Figure \ref{fig:uav}.
The vehicle has four 205 mm rotors attached to a UAV frame.
Two microphones are centered between the front rotors and positioned normal to the wall.
The microphones have a spacing of 15 mm, and the rear mic is 350 mm from the center of the front propellers.
The microphones are connected to a Teensy 4.0 microcontroller, which computes distance estimates in real time from the streaming audio signal.

To collect a distance ground truth, an Adafruit VL53L0X optical time-of-flight ranging sensor is positioned 175 mm inboard of the microphones. The sensor is connected to an Adafruit Feather HUZZAH microcontroller for logging and communication.
In all tests, wall distance is measured from the center of the microphone array.

\subsection{Actuating Arm Setup}\label{sec:arm}

For all experiments, the UAV is mounted to a HEBI 6DoF actuating arm, as seen in Figure \ref{fig:arm}.
In the following tests, the arm moves the vehicle along an axis normal to the wall, keeping the vehicle level.
Arm motion is commanded using the output of a feedback control algorithm using onboard distance estimates.

\subsection{Distance Estimation Algorithms}

We estimate distance with two methods described in \cite{Cal21} and Algorithm \ref{alg:bf}, which we will denote as \textit{channel} and \textit{beam}.
In addition, we use an optical \textit{rangefinder} to provide a distance ground truth for both open- and closed-loop control tests.
We correct for the position offset between the rangefinder and origin (center of the front microphone) to provide a ground truth for evaluating our distance estimation algorithms.
Additionally, we use the rangefinder to demonstrate baseline performance of our closed-loop control system with ``ideal'' distance measurements.

\begin{figure}
  \centering
  \includegraphics[width=\linewidth]{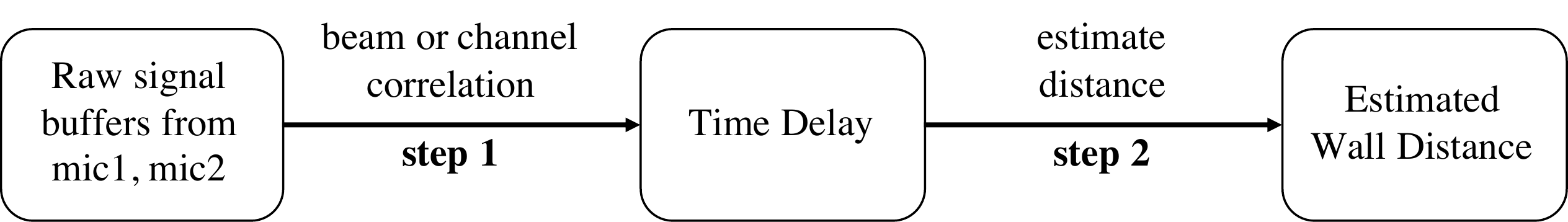}
  \caption{Acoustic distance estimation occurs in two high-level steps: (a) finding the delay timeseries, and (b) estimating distance from that timeseries. In this work, we investigate step 1.}
  \label{fig:two-step-dist}
\end{figure}

Beam and channel correlation approaches both estimate the wall distance analytically from the acoustic signal (Figure \ref{fig:two-step-dist}). Distance estimation occurs in two steps:

\begin{enumerate}
  \item \textit{Time delay series.}
    First, the raw signal buffers are processed using either the channel correlation \cite{Cal21} or beamforming algorithm, resulting in a delay timeseries; an example can be seen in Results in Figure \ref{fig:static-iffts}.
    The time delay corresponding to the reflected wall noise at each timestep can be seen visually as the bin with the highest magnitude.
  \item \textit{Distance estimation.}
    Next, a simple estimation algorithm extracts this signal and converts the time delay to an estimated wall distance.
    As this step is not the focus of this work, we use the same simple algorithm for both the beam and channel methods: we select the top three time delay bins, choose the bin closest to the previous bin, find the corresponding round-trip distance, and half that to get the wall distance:

    \begin{equation}
      d = t_d * c / 2
    \end{equation}

    where $d$ is the estimated wall distance, $t_d$ is the time delay, and $c\approx 343$ m/s is the speed of sound.

\end{enumerate}

Both channel and beamforming distance estimates are computed fully onboard the Teensy at a rate of 10 Hz.

\subsection{Control System}

In closed-loop control experiments, the arm movement is commanded using a correction $x_c$ proportional to the difference between current (estimated) and target wall distances, $x_p:=d$ and $x_t$:

\begin{equation}
  x_c = K_p (x_t - x_p)
\end{equation}

where $K_p$ is the proportional gain.

The control loop runs at 10Hz, matching the distance estimation frequency.

\subsection{Experiments}
\subsubsection{Open-Loop Control Experiments}

We compare the onboard distance estimates from the beam- and channel-based correlation algorithms while the arm moved in a oscillating pattern with respect to the wall.
All predictions and intermediate values are computed onboard the Teensy and streamed in real time.

\subsubsection{Closed-Loop Control Experiments}

\begin{figure}
  \centering
  \includegraphics{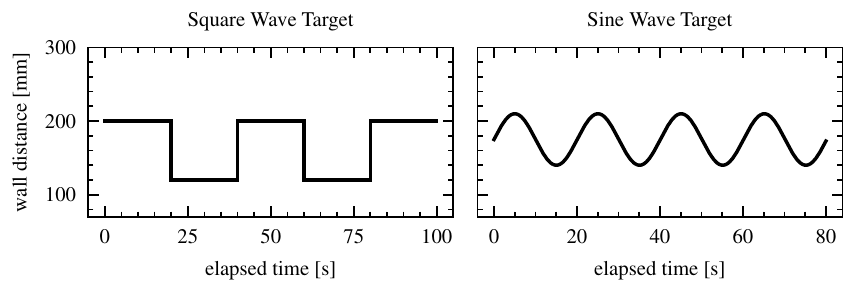}
  \caption{Commanded distance for square (left) and sine (right) wave tests}
  \label{fig:command}
\end{figure}

We test closed-loop control by sending each distance estimate and target to a proportional controller on the HEBI arm as described previously.
We conduct two tests:

\begin{enumerate}
  \item \textit{Square Wave Test.} The vehicle is commanded to move between two target wall distances: 120 mm and 200 mm. The test runs for 2.5 periods, with 20 s at each distance for a total test time of 100s (Figure \ref{fig:command}, left).
  \item \textit{Sine Wave Test.} The vehicle is commanded to oscillate between two distances. We set the target distance with a sin wave centered at 175 mm with an amplitude of 35 mm. The test runs for 4 periods, each lasting 20 s for a total test time of 80 s (Figure \ref{fig:command}, right).
\end{enumerate}

We run each test three times for each wall distance method: rangefinder (baseline), beam, and channel.
We compare the position trajectories for each method as well as the distance estimate distributions compared with the ground truth. For each test, we compute steady-state error and variance vs. the commanded distance. In addition, for the square wave test, we compute the average rise time (time required to get from 10\% to 90\% of a commanded distance), and average overshoot.

\subsubsection{Distance Test}

Finally, we test the distance limit of both methods by comparing onboard estimates at different wall distances ranging from 4 cm to 0.7 m.

\section{Results}\label{sec:results}

\begin{figure}[b]
  \centering
  \includegraphics{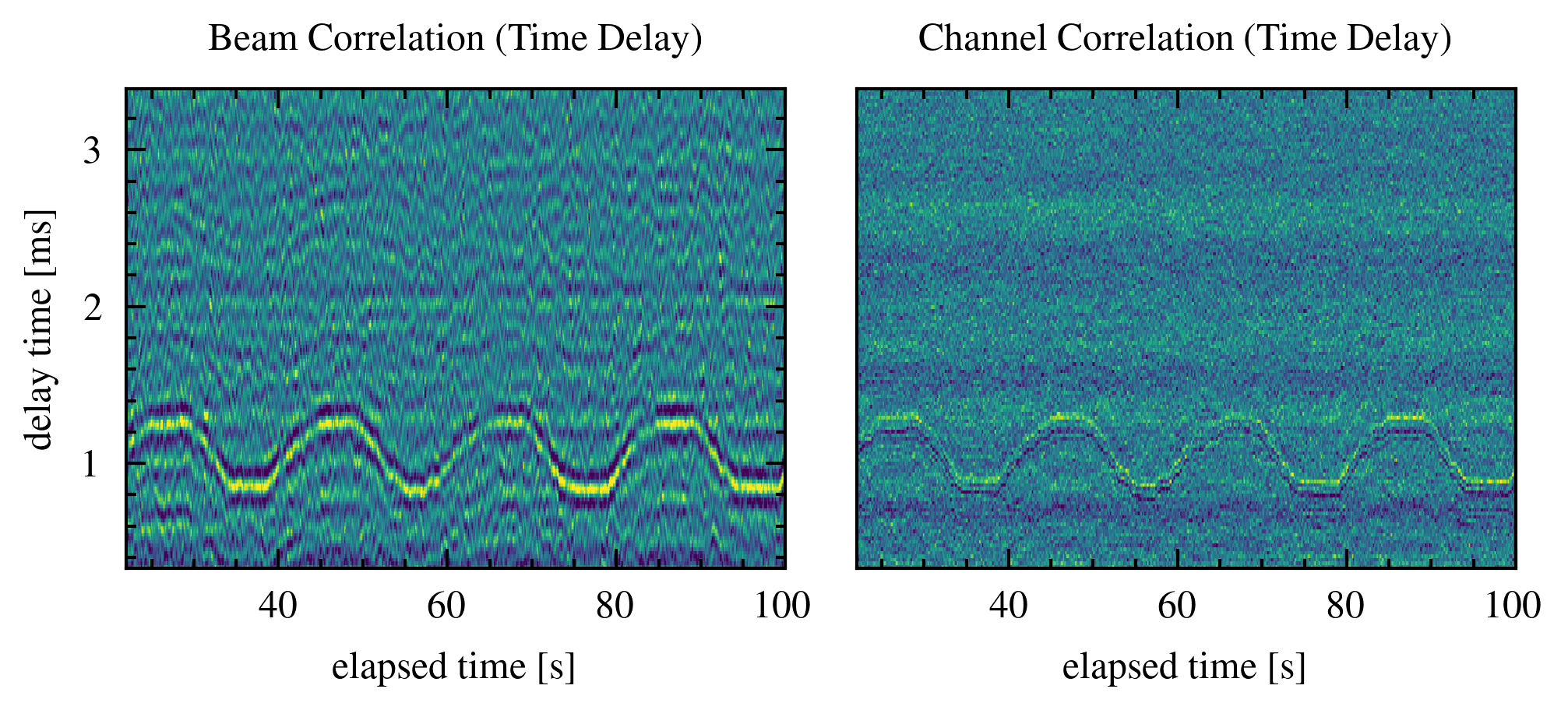}
  \caption{Beam vs. channel delay timeseries for an oscillating arm motion.}
  \label{fig:static-iffts}
\end{figure}

\subsection{Open-Loop Control}\label{sec:open-loop}

\subsubsection{Oscillation Test}

\begin{figure}[ht]
  \centering
  \includegraphics{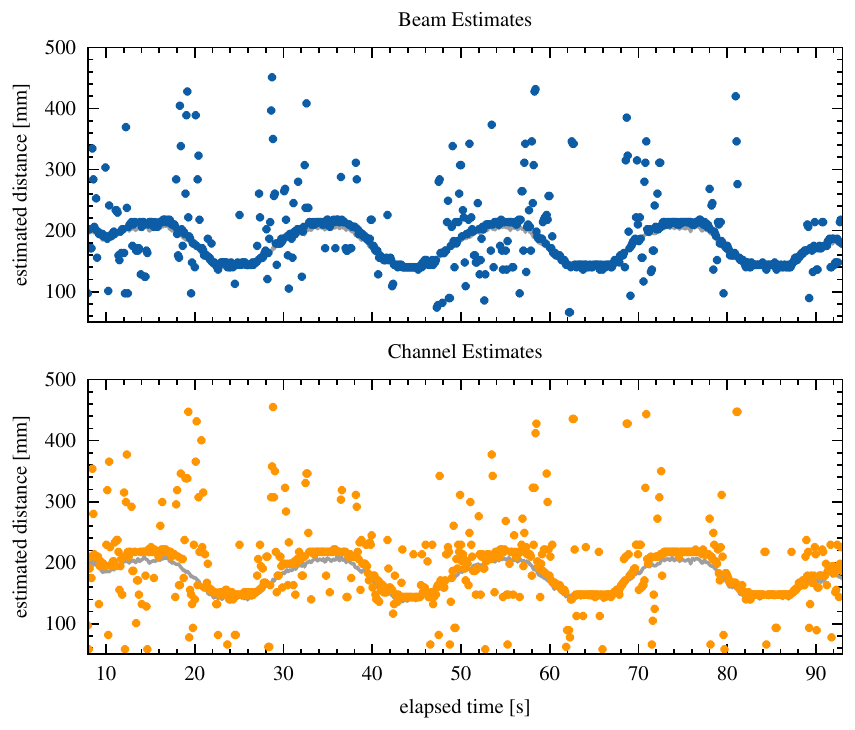}
  \caption{Onboard predictions for sinusoidal motion using the beamforming and channel-based correlation algorithms. The top figure shows predictions at each timestep plotted against the rangefinder ground truth, and the bottom figure shows the associated error plots. Both methods show increasing error as wall distance increases with consistently lower error for beamforming.}
  \label{fig:static-preds}
\end{figure}

We begin by comparing onboard predictions for beam vs. channel during an arm oscillation test.
Qualitatively, the beamforming method produces a much clearer signal in the delay timeseries than the channel correlation method (Figure \ref{fig:static-iffts})
as well as more accurate and precise distance estimates (Figure \ref{fig:static-preds}).
Distance estimates extracted from these timeseries demonstrate that the beamforming method produces more accurate and precise predictions than the channel correlation method (Figure \ref{fig:static-preds}).
Quantitatively, beam outperforms channel while maintaining the same prediction rate (Table \ref{tab:open}).

\begin{table}[ht]
  \footnotesize
  \begin{tabular}{@{}lllll@{}}
    \toprule
    Method            & Avg. Error & Std.Dev. & Max Error & Output Freq. \\ \midrule
    Cross-correlation & 35.0 mm                & 61.2 mm                     & 291.4 mm           & \textbf{10 Hz}   \\
    Beamforming       & \textbf{27.3 mm}       & \textbf{53.3 mm}            & \textbf{283.1 mm}  & \textbf{10 Hz}
    \\\bottomrule
  \end{tabular}
  \caption{Open-Loop Control Error}
  \label{tab:open}
\end{table}

Both methods produce estimates at approximately the same frequency (10 Hz), which matches the speed of our control loop.

\subsection{Closed-Loop Control}\label{sec:closed-loop}

After open-loop analysis, we conducted several closed-loop control tests, which feedback beam and channel distance estimates in the arm control loop, as described previously.
Both methods were successful at controlling square and sine wave motion, though the beam method was consistently more accurate than channel correlation.

\subsubsection{Square Wave}

\begin{figure}[t]
  \centering
  \includegraphics{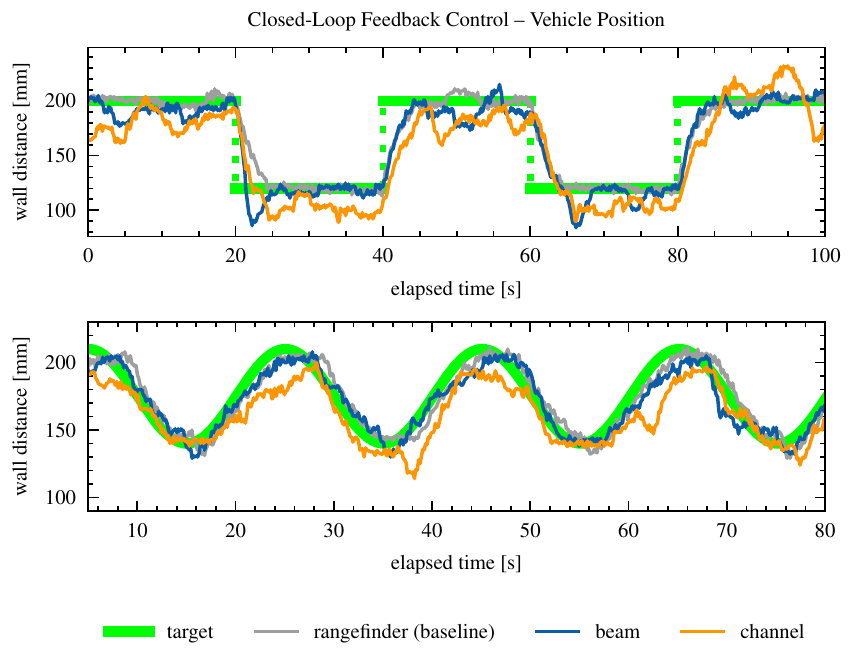}
  \caption{Positions during closed-loop control using distance predictions from each algorithm (rangefinder ground truth, acoustic beam correlation, and acoustic channel correlation.)}
  \label{fig:active}
\end{figure}

Results from the square wave tests can be seen in Figure \ref{fig:active}, showing the rangefinder-measured wall distance using the specified feedback (baseline control using rangefinder measurements; beam; and channel) plotted against the commanded distance.
Both the beam and channel methods tend to overshoot when moving closer to the wall, which may be due to an external moment on the HEBI arm caused by the vehicle weight; however, the beamforming method is able to move back to the target position more effectively after this overshoot.
Through several tests, both methods also lead to the vehicle undershooting the target distance when moving away from the wall.
We believe this is due to a skew in outliers towards higher values, which can be seen in Figure \ref{fig:active-preds-sin} (left).

The beam correlation control was twice as accurate as the channel correlation control during steady state, had a 1.5 s faster rise time, and had a smaller incidence of large outliers (Table \ref{tab:active-square}).
While beam experienced more significant overshooting in motion towards the wall, it usually recovered to the commanded distance quickly;
in contrast, channel struggled to maintain the target distance, often reaching a steady state several centimeters off the target (Figure \ref{fig:active}).

\begin{table}[]
  \scriptsize
    \begin{tabular}{@{}rllll@{}}
    \toprule
    Method&
      \begin{tabular}[c]{@{}l@{}}Steady-State\\Error v. Target\end{tabular}&
      \begin{tabular}[c]{@{}l@{}}Steady-State\\Std.Dev.\end{tabular}&
      Rise Time&
      Overshoot \%\\ \midrule
    \begin{tabular}[c]{@{}l@{}}\textit{Rangefinder}\\\textit{ (baseline)}\end{tabular}&
      \textit{2.8 mm} &
      \textit{2.3 mm} &
      \textit{3.77 s} &
      \textit{n/a} \\
    Channel& 34.9 mm & 40.3 mm & 6.11 s & \textbf{32.39\%} \\
    Beam& \textbf{19.7 mm} & \textbf{34.4 mm} & \textbf{4.59 s} & 43.06\%
    \\\bottomrule
    \end{tabular}
    \caption{Closed-Loop Feedback Control Error}
    \label{tab:active-square}
\end{table}

\begin{table}[b]
  \footnotesize
  \centering
  \begin{tabular}{@{}rlll@{}}
    \toprule
    Method &
      \begin{tabular}[c]{@{}l@{}}Avg Error\\ v. Target\end{tabular} &
      Std.Dev. &
      \begin{tabular}[c]{@{}l@{}}Max Error\\ v. Target\end{tabular} \\ \midrule
    \begin{tabular}[c]{@{}l@{}}Rangefinder\\ (baseline)\end{tabular} &
      \textit{8.4 mm} &
      \textit{4.0 mm} &
      \textit{22.0 mm} \\
    Channel &
      13.1 mm &
      10.3 mm &
      51.7 mm \\
    Beam &
      \textbf{7.7 mm} &
      \textbf{5.0 mm} &
      \textbf{25.1 mm}
    \\\bottomrule
  \end{tabular}
    \caption{Error metrics for closed-loop sine wave test}
    \label{tab:active-sin}
\end{table}

\subsubsection{Sine Wave}

Similarly, during the sine wave test, the vehicle oscillates between two commanded distances.
This test encourages continuous motion, resulting in more background noise from the arm actuators.
Because of this, the results indicate how a vehicle might perform in noisier conditions.

Results can be seen in Figure \ref{fig:active} (bottom), which shows the position of each method plotted against the commanded distance.
While the smooth motion of the sine wave facilitates less overshooting compared to the square wave test, the channel method struggles with this test's increased background noise.
Significant position errors can be seen at 38 s and 63 s, and the channel predictions through the full test are much less precise than the corresponding beam predictions (Table \ref{tab:active-sin}).
Note that error is computed with respect to the commanded distance.

\begin{figure}
  \centering
  \includegraphics[width=\linewidth]{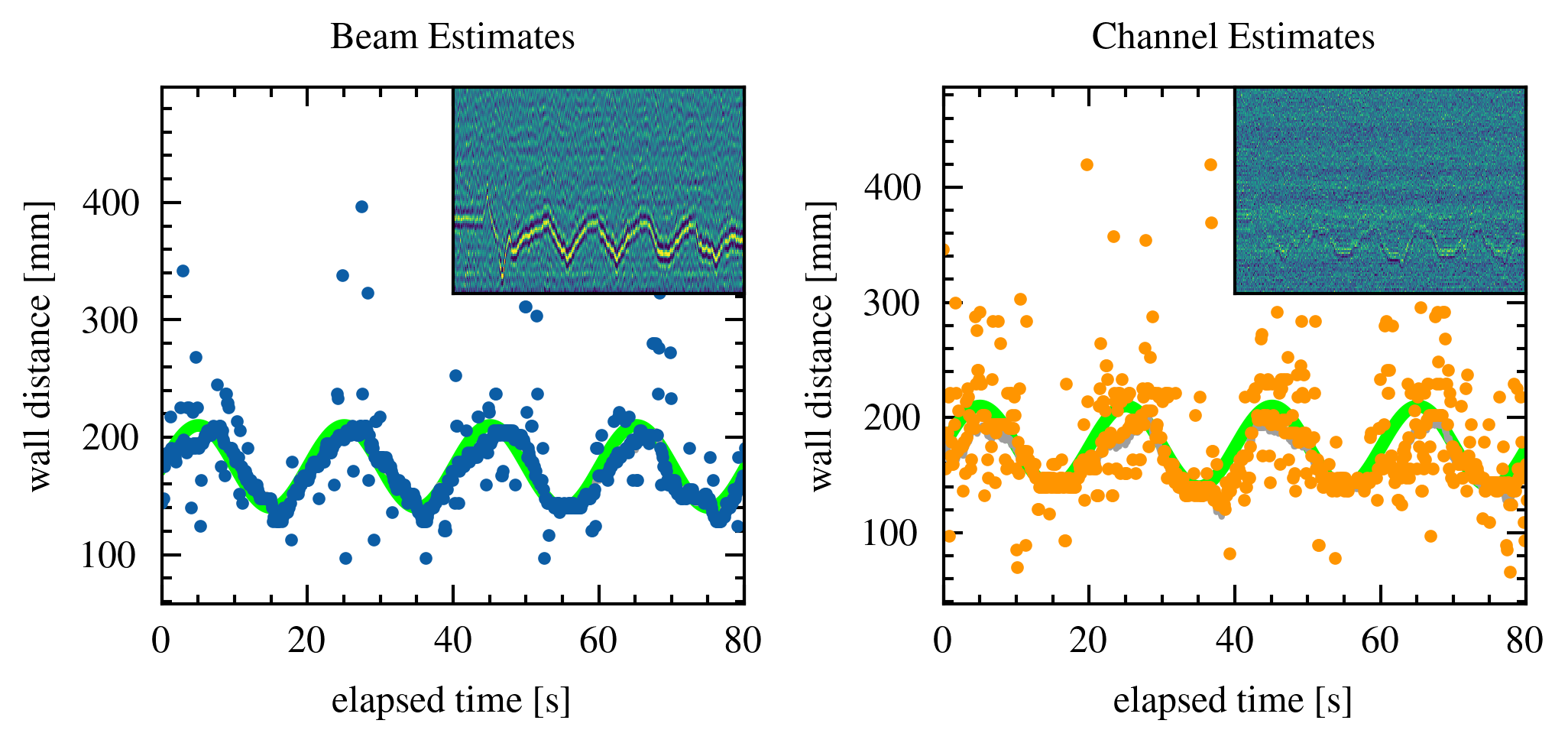}
  \caption{Distance estimates for each closed-loop control run showing the commanded distance (green line) with overlayed estimates (dots). The delay timeseries plots are inset for qualitative comparison.}
  \label{fig:active-preds-sin}
\end{figure}

This trend can be seen even more clearly when we compare distance estimates between the two methods (Figure \ref{fig:active-preds-sin}).
The higher noise in the channel approach compared with beam is consistently more pronounced in closed-loop control tests than in the previous open-loop tests.
We believe this is due to a vicious control cycle: the noisier predictions result in more arm movement, which results in higher ambient noise and even noisier predictions.
While this cycle would likely be less pronounced in a free-flight situation, we believe it would still be present.
In contrast, the beamforming predictions match the noise from open loop test results.

We repeated these closed-loop feedback tests in several more challenging conditions, including various ambient noise conditions, multiple nearby reflectors, and against a rough rock wall, and the beam approach matched its performance in more controlled conditions.
Future work will more thoroughly investigate robustness to complex environments and characterize failure modes of this approach.

\subsection{Distance Test}\label{sec:dist}

\begin{figure}
  \centering
  \includegraphics[width=\linewidth]{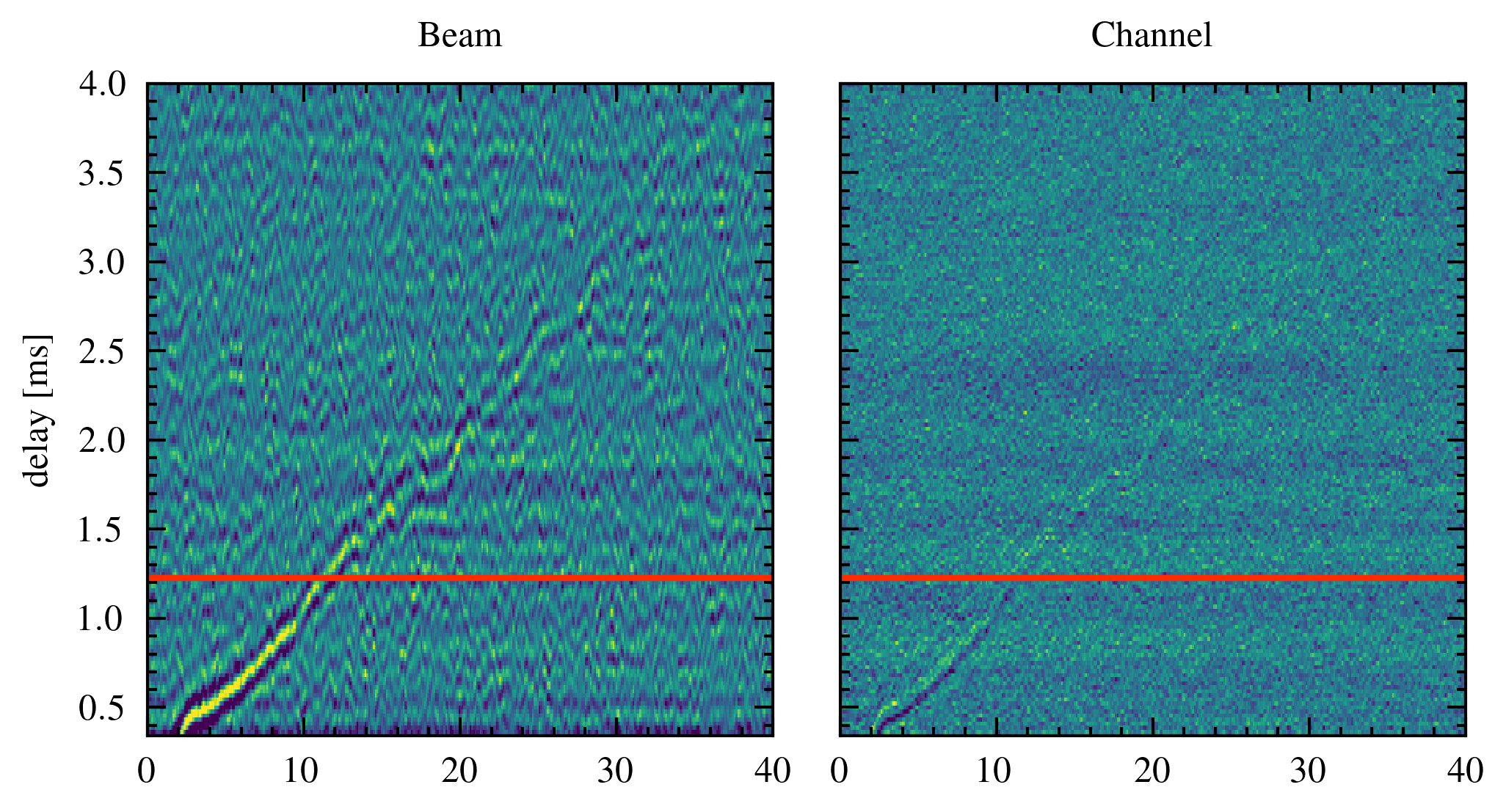}\\
  \includegraphics[width=\linewidth]{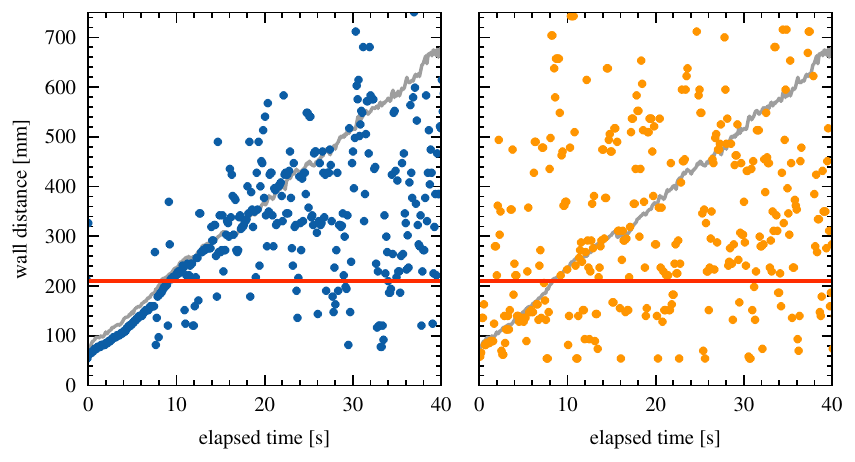}
  \caption{Delay time for each acoustic estimation method at distances ranging from 4 cm to 0.7 m. The red line indicates the limit of our current closed-loop tests.}
  \label{fig:dist}
\end{figure}

A useful algorithm for obstacle detection should be able to detect obstacles at a sufficient distance for the UAV to react.
Figure \ref{fig:dist} shows time delay and distance predictions for both algorithms between 4 cm and 0.7 m, the maximum range of the optical rangefinder with this vehicle geometry.

In both the channel and beam delay timeseries,
the signal attenuates as the wall distance increases,
As before, the beam-based correlation signal appears clearer than the channel-based signal.
Using the simple distance estimation algorithm, the channel is accurate enough for closed-loop control up to approximately 200 mm, similar to \cite{Cal21}.
The beam method maintains a similar error and variance more than 2x further (up to nearly 0.5 m.)
Qualitatively, the peak can be seen clearly for time delays approaching 4 ms with beam, corresponding to a 0.7 m wall distance.

While our closed-loop control tests are mostly conducted in the 100--200 mm range,
these results are encouraging for the beamforming method being effective for active control at much further wall distances, particularly as the distance estimation algorithm from Figure \ref{fig:two-step-dist} is improved.

\section{Conclusion}\label{sec:discussion}

In this work, we present a method of estimating object-relative distance from a quad-rotor UAV by applying acoustic beamforming to the implicit noise from the vehicle propulsion system. We show high estimation accuracy in an open-loop setting, implement this approach onboard the vehicle for real-time prediction, and demonstrate successful closed-loop feedback control of the arm-mounted UAV. The beam-based correlation approach consistently outperforms a channel-based correlation approach in simulation, open-loop analysis, and finally closed-loop onboard feedback control experiments.
Furthermore, open-loop analysis at varying wall distances show that the beamforming signal is strong at wall distances approaching 0.7 m, resulting in accurate estimates at 2x further distances.
This is promising for eventual free-flight: detecting a wall at a further distance will make crashes less likely, make vehicles more maneuverable, and facilitate more complex acoustic sensing and mapping tasks.

An immediate extension of this work is applying acoustic beamforming approach to object-relative orientation estimation for coarse 2D sensing.
We also plan to better characterize the limits of this approach, investigating robustness to different environmental and operational settings, wall distances, and vehicle configurations.
Finally, we intend to test this method in a free-flight setting, beginning with open-loop analysis and leading to closed-loop feedback control of a flying UAV.










\bibliographystyle{IEEEtran}
\bibliography{root.bib}

\end{document}